\documentclass[conference]{IEEEtran}
\IEEEoverridecommandlockouts

\usepackage{cite}
\usepackage{amsmath,amssymb,amsfonts}
\usepackage{algorithmic}
\usepackage{graphicx}
\usepackage{textcomp}
\usepackage{colortbl}
\usepackage{booktabs}
\usepackage{subcaption}
 
\usepackage[table,xcdraw]{xcolor}
\def\BibTeX{{\rm B\kern-.05em{\sc i\kern-.025em b}\kern-.08em
    T\kern-.1667em\lower.7ex\hbox{E}\kern-.125emX}}
\begin{document}

\title{Congenital Heart Disease recognition using Deep Learning/Transformer models \\
}

\author{\IEEEauthorblockN{Aidar Amangeldi}
\IEEEauthorblockA{\textit{MSc Data Science, NU} \\
aidar.amangeldi@nu.edu.kz}
\and
\IEEEauthorblockN{Vladislav Yarovenko}
\IEEEauthorblockA{\textit{PhD Computer Science, NU} \\
vladislav.yarovenko@nu.edu.kz}
\and
\IEEEauthorblockN{Angsar Taigonyrov}
\IEEEauthorblockA{\textit{PhD Computer Science, NU} \\
angsar.taigonyrov@nu.edu.kz}
}

\maketitle

\begin{abstract}
Congenital Heart Disease (CHD) remains a leading cause of infant morbidity and mortality, yet non‐invasive screening methods often yield false negatives. It is believed that Deep Learning models, with their advantage of automatically extracting features, can aid doctors in detecting the CHD more effectively. Thus, we investigated the possibility of using dual‐modality (sound+image) deep learning methods in this domain. 73.9\% accuracy on ZCHSound dataset and 80.72\% on DICOM Chest X-ray dataset. 
\end{abstract}


\section{Introduction}
Congenital Heart Disease (CHD) is one of the most common heart diseases accounting for approximately 33\% of major birth defects and 30-50\% of infant mortality cases \cite{10.1093/eurheartj/ehaa874,zhang2015diagnostic}. While universal newborn screenings and fetal echocardiography are effective in CHD detection, there is still a chance of a false negative result. Recent studies have shown the potential of Computer Vision (CV) techniques, such as Convolutional Neural Networks (CNNs) to improve  CHD diagnosis and prediction \cite{khan2025rolemachinelearningcongenital}. Common data collection techniques for this task include X-ray and auscultation, which are usually used independently \cite{sharifi2025detection,10384868}.\par

This project aims to develop separate architectures that learn CHD characteristics from X-rays and visual representations of heart sound recordings, potentially combining these findings for more accurate diagnoses. This method will focus on detecting universal features that are present in selected data representations, improving model generalization. \par

The contributions of this project are as follows:
\begin{itemize}
    \item Develop and train CNN architectures to learn CHD features from X-ray and phonocardiogram (PCG) data separately.
    \item Evaluate resulting Deep Learning models’ ability to detect and classify CHD with standardized performance metrics.
    \item Attempt to implement a fusion mechanism that combines predictions from both models.
\end{itemize}

\newpage
\section{Related Works}
This section explores existing work in CHD recognition and summarizes similar research works.
\subsection{Spectrogram based approaches}
Yang et al. \cite{yang2023assisting} proposed a Transformer-based model for classification of heart sound signals. It is aimed at assisting the diagnosis of heart diseases located at its valves. Traditional binary classification methods are limited in their ability to assess minor differences in heart sounds that reflect varying heart conditions. The model (10-fold CV) yields ~100\% in all metrics: accuracy, sensitivity, specificity, precision, and F1.
Jia et al. \cite{inproceedings} proposed a method that uses wavelet decomposition and Normalized Average Shannon Energy for feature extraction. A novel fuzzy neural network with structure learning is then employed for heart sound classification.
Zhang et al. \cite{ZHANG201720} proposed a method called a scaled spectrogram and partial least squares regression (PLSR) working on a PASCAL dataset. 
Hamidi et al. \cite{article} proposed two feature extraction methods: a so-called curve fitting feature extraction and MFCC+fractal features. Classification was performed using the kNN algorithm.

\subsection{X-ray based approaches}
Zhixin et al. \cite{zhixin2024chd} utilized ResNet18 as the base model. The model was pre-trained on the ImageNet dataset using transfer learning. Data augmentation techniques like randomized tiling, color inversion and mirroring were applied to reduce overfitting. In this way, the model performance was evaluated both overall and compartment-wise. Lastly, Class Activation Mapping (CAM) was used to generate heatmaps on X-ray images which helped experts to interpret the results. 
Fedchenko et al. \cite{10.1093/eurheartj/ehaa874} used transfer learning with a pretrained Inception-v3, which was initially trained on the ImageNet dataset. The final layer of the base model was replaced with a global average pooling layer and a fully connected layer to predict the pulmonary-to-systemic flow ratio.
Alternatively, Jiang et al. \cite{inbook} utilized a relatively new approach called Vision Transformers. They proposed a lightweight hybrid model called FlashViT. By implementing an unsupervised homogenous pre-training strategy they utilized less than 1\% of class-agnostic medical images. Compared to conventional approaches, ViT appears to be more efficient in limited data and poor generalization cases.

\section{Dataset description}

The research utilizes two independent datasets: ZCHSound heart sound recordings and DICOM chest x-ray files. The mentioned datasets provide supplementary information for deep learning model training and evaluation. Some researchers stated that there are around thirty-five types of CHDs \cite{shabana2020genetic}. However, most of them are not fully explored by medical practitioners. Therefore, in this project, only four types of CHDs are used based on utilized datasets in the project: ASD, VSD, PDA, PFO.\par

\subsubsection{ZCHSound}

ZCHSound was collected at the Children’s Hospital of Zhejiang University where the auscultations were gathered utilizing a stethoscope with an 8000 Hz frequency range \cite{khan2025rolemachinelearningcongenital}. 
The heart sound dataset consists of 941 participants and 941 audio recordings, each approximately 20 seconds long, totaling over 5 hours in duration. This dataset includes 473 females (50.27\%) and 468 males (49.73\%), with 533 participants without heart disease acting as a control group, while the other patients were diagnosed with different heart disease types like atrial septal defect (ASD) (119 cases), patent ductus arteriosus (PDA) (32 cases), patent foramen ovale (PFO) (70 cases), and ventricular septal defect (VSD) (187 cases) [6]. The dataset provides the labeled examples of both normal and abnormal heart sounds but presents class imbalance.\par

\subsubsection{DICOM}
\begin{figure}[ht]
  \centerline{\includegraphics[width=0.25\textwidth]{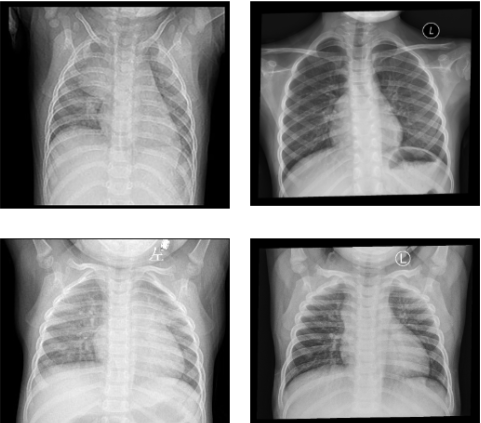}}
  \caption{DICOM: ASD, Normal, PDA and VSD types of CHD.}
  \label{fig:cxr}
  \vskip-3mm
\end{figure}

DICOM \cite{zhixin2024chd} is chest x-ray dataset images collected from Qingdao Women and Children Hospital between 2021 and 2022. Originally the storage format of the files was DICOM. As it contains sensitive patient information such as age, name, and execution time, the format of the x-ray files was converted to JPEG by the de-identification program to ensure patient privacy.
The dataset consists of 828 chest x-ray images from 473 female and 355 male children, with an average age of 6.7 years categorized into 4 classes: ASD (194 cases), VSD (210 cases), PDA (216 cases), and normal control group (218 cases). Figure \ref{fig:cxr} demonstrates the chest x-ray images by the classes.\par

\newpage
\section{Methodology}
\subsection{Audio Late Fusion}

\begin{figure}[ht] 
  \centerline{\includegraphics[width=0.49\textwidth]{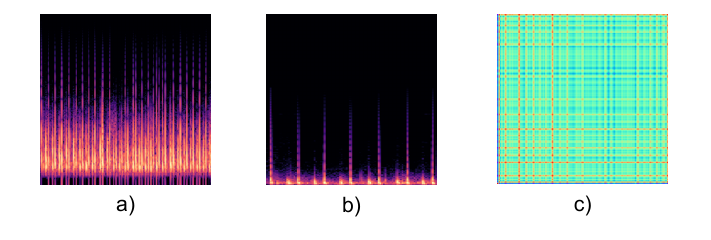}}
  \caption{STFT (a), Mel (b), and GAF (c) Transformations of Audio Data.}
  \label{fig:audio_vis}
  \vskip-3mm
\end{figure}

Previous research in CHD prediction shows that the use of 2-dimensional Convolutional Neural Networks can improve the prediction accuracy. 
To efficiently use these models, the audio data from the ZCHSound dataset has to be converted into a 2D representation. 
Currently, several algorithms can be used for that:
\begin{itemize}
    \item \textbf{Short-Time Fourier Transform (STFT)}. Breaks the audio signal into short time windows and computes the Fourier transform for each window. This representation makes it possible to observe how the frequency evolves over time, which is important for detecting dynamic heartbeat features. 
    \item \textbf{Mel Spectrogram}. Uses an STFT frequency axis and applies a perceptually-motivated scale to it, amplifying frequencies that are more relevant to human hearing. Simultaneously, it highlights low-frequency patterns and reduces noise, which is helpful for medical audio analysis.
    \item \textbf{Gramian Angular Fields (GAF)}. Encodes angular correlations in polar coordinates. This preserves the dynamics of heartbeat signals and captures variations in rhythm and amplitudes.  
\end{itemize}
Using these transformations, it is possible to create and visualize different 2D representations of the heart audio recordings, making it a suitable modality for CNNs. 
Figure \ref{fig:audio_vis} illustrates the difference between different transformations. \par

The next step is to test the performance of several CNN architectures on each data representation.
For this, we selected ResNet-50V2, EfficientNetB0, and InceptionV3, as these models have shown their successful results in CHD prediction. 
These models will be tested for each data representation, and the best architecture for each one will be selected. \par

\begin{figure*}
  \centerline{\includegraphics[width=0.7\textwidth]{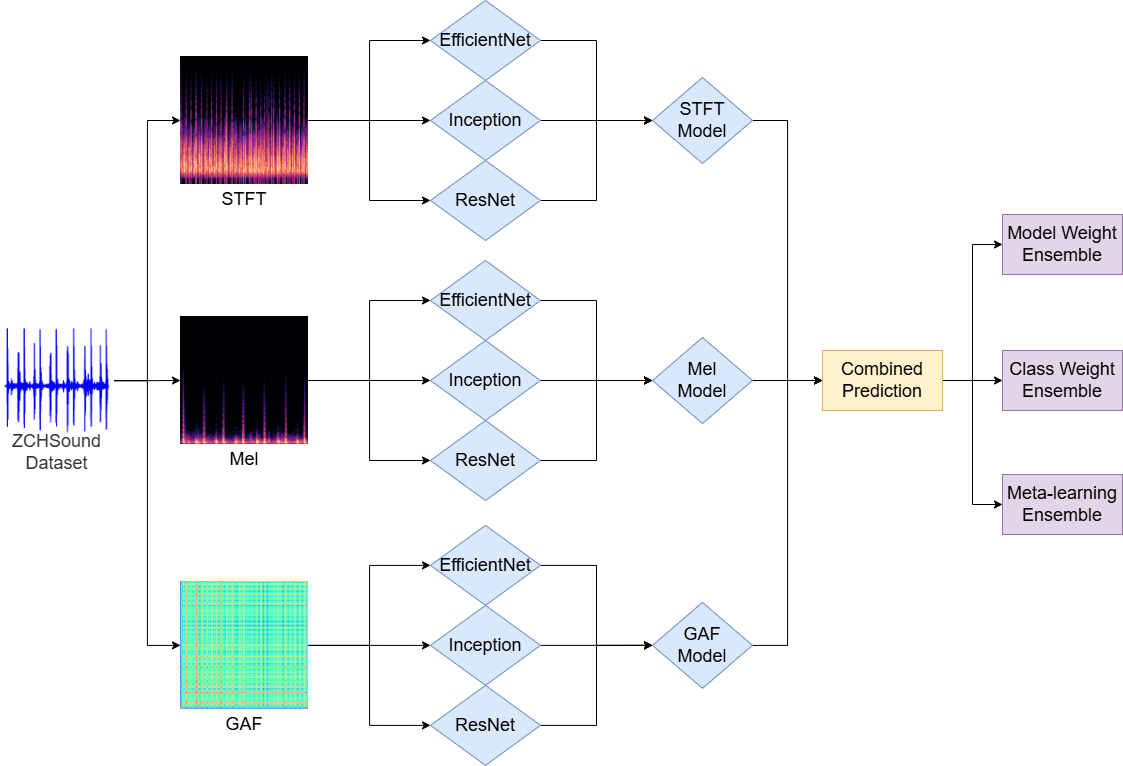}}
  \caption{An Overview of the Audio Processing Methodology.}
  \label{fig:audio_method}
  \vskip-3mm
\end{figure*}

Finally, the classification results of these models will be combined using several late fusion approaches. 
We test two types of weights: the validation accuracy of each model, and the class F1 scores for each class of each model. 
Additionally, the meta-ensemble approach that uses Logistic Regression model to learn how to optimally combine the predictions from all base models.
These three approaches will be compared in terms of their effect on the accuracy and F1 scores, and whether they improve the results of base models. Figure \ref{fig:audio_method} illustrates the complete framework of the methodology. \par

\begin{figure*}
  \centerline{\includegraphics[width=0.7\textwidth]{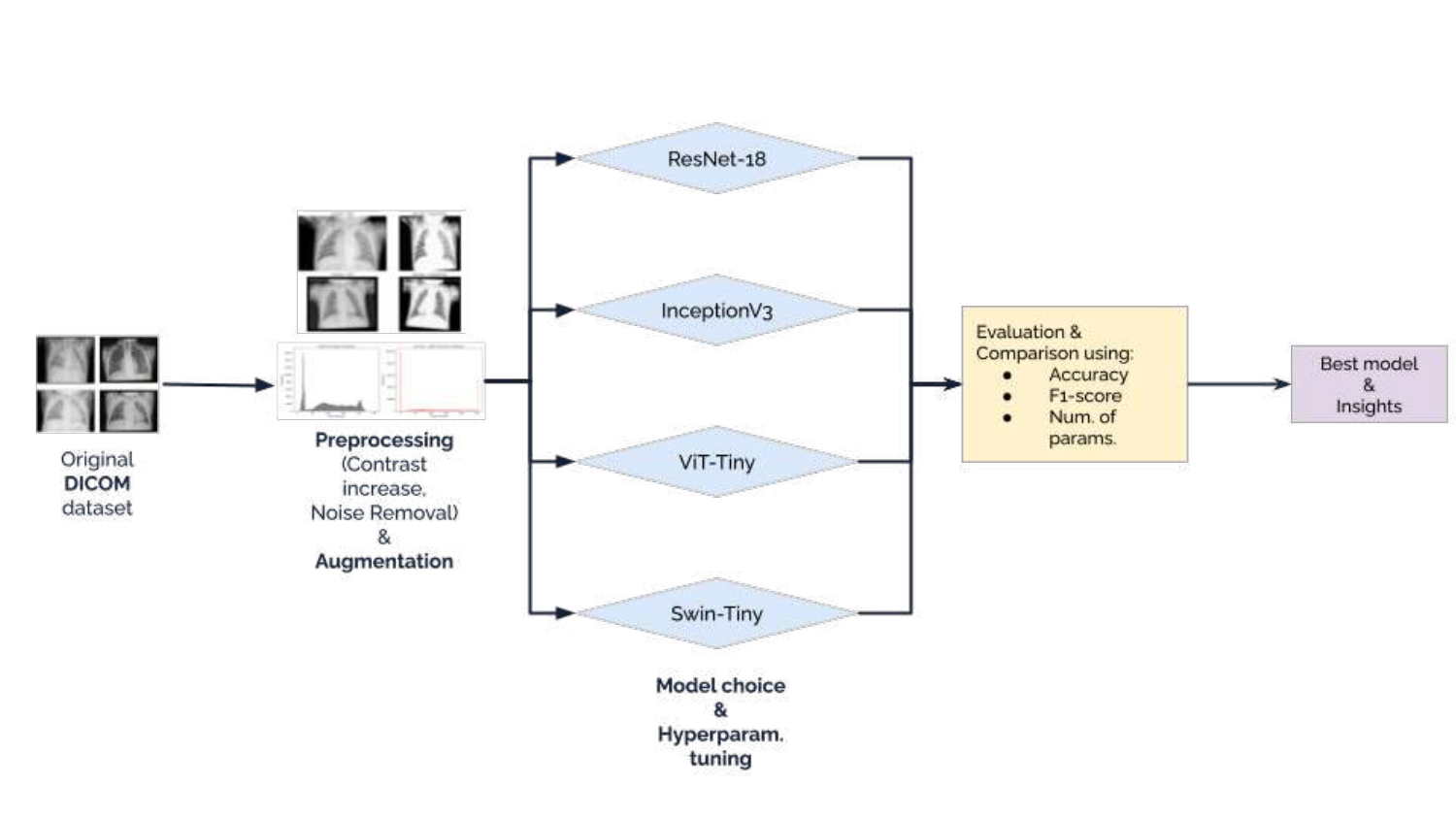}}
  \caption{An Overview of the Image Processing Methodology.}
  \label{fig:dicom-method}
  \vskip-3mm
\end{figure*}

\subsection{Classifiication models for DICOM CHD-CXR images dataset}
As it is crucial to have clear and high quality X-ray images for an accurate diagnosis in medical imaging, we applied image enhancement and blurring techniques in the preprocessing stage.

\textbf{Gaussian Blurring Filter.} To address the noise in the image, Gaussian Blurring Filter (GBF) was used that effectively reduced unwanted variations and preserved crucial anatomical details. GBF is a widely used image pre-processing technique in medical imaging, especially for denoising and detail preservation. It was first formalized in the context of computer vision by Marr and Hildreth \cite{marr1980theory} who proposed the Gaussian filter as a key step in edge detection through the Laplacian of Gaussian (LoG). In general, the Gaussian filter is applied to smooth images and suppress noise using a convolution with a Gaussian kernel, defined as:

\[
G(x, y) = \frac{1}{2\pi\sigma^2} \exp\left(-\frac{x^2 + y^2}{2\sigma^2}\right)
\]

where $\sigma$ controls the degree of blurring (in our case 0.8). The Gaussian kernel effectively attenuates high-frequency components (noise) while preserving low-frequency structures, which is vital for highlighting anatomical features in medical images. Several prior studies have demonstrated the effectiveness of Gaussian-based preprocessing for X-ray classification. 
Devi et al. \cite{devi2023gaussian} utilized Gaussian blurring in conjunction with HSV color space transformations to effectively segment and classify Leukemia cells from biopsy images. Their approach achieved high accuracy on both private and public datasets. Similarly, Chowdhury et al. \cite{chowdhury2021can} employed Gaussian blurring in the preprocessing pipeline to denoise images and boost model performance in pneumonia classification tasks.

This step enhanced the quality of the input data by reducing irrelevant noise and improving the clarity of soft tissues. To be precise, we gained around 2-3\% increase in accuracy. For the fairness, it should be said that due to time constraints we were not able to analyze the histograms of images for the presence of specific types of noise and we assumed that GBF works good in most of the cases. Nevertheless, in other papers authors have seen even higher performance boost.

\textbf{Contrast Enhancement.} Next, the contrast enhancement was applied (contrast factor = 1.8) to improve the visibility of features in medical images by adjusting the intensity distribution, making subtle anatomical structures more distinguishable. Histogram equalization method was used. It redistributes pixel intensities to span the full dynamic range. The transformation function can be defined as:

\[
s = T(r) = \int_0^r p_r(w) \, dw
\]

where \( p_r(w) \) is the probability density function of the pixel intensities in the original image, and \( T(r) \) maps the original intensity \( r \) to the new intensity \( s \).

Previous studies have shown that contrast enhancement can significantly improve classification performance in medical imaging. For instance, Tang et al. \cite{tang2020automated} applied contrast-limited adaptive histogram equalization (CLAHE) to improve COVID-19 detection from chest X-rays. Likewise, Pasa et al. \cite{pasa2021efficient} used global contrast enhancement to increase image clarity, which improved deep neural network classification of pneumonia and other thoracic diseases.

By increasing the contrast between bones, soft tissues, and potential disease markers, the classifier could focus on more informative regions of the image. The accuracy gain (along with GBF) was 6-7\% and a better separability between classes (based on F1 score). Like in noise reduction, there is a way for improvement: one can test various contrast factors to get the most optimal one. Again, some average value was picked that we assume works decent in most of the cases.

\textbf{Augmentation.} Further our training set was augmented by using small rotations (±5°), horizontal flips, and minor brightness and contrast adjustments. These techniques improved the diversity of the dataset. Then all images were uniformly resized to 224×224 pixels and normalized using standard ImageNet statistics that ensures consistency with the pretraining conditions of our models.

\textbf{Training loop.}
The models were trained using an 80/20 train/validation split with batch sizes of 32 for both loaders. Training proceeded for up to 100 epochs. Early stopping with a patience of 7-10 epochs monitored the validation accuracy to halt training once no improvement was observed for seven consecutive epochs, thereby preventing overfitting and ensuring stable convergence.

\section{Results and Discussion}

\subsection{Audio Late Fusion}

The first step is to evaluate all selected model architectures to select the best performing one for each data representation.
Table \ref{tab:zch_results} shows that on average, the STFT representation has the highest testing accuracy, achieving at least 71.1\%. 
Based on the overall results, we choose ResNet for STFT, and Inception for Mel and GAF, as they provide the best testing accuracy for each data representation. \par
\begin{table*}[ht]
  \centering
  \begin{subtable}[t]{0.58\textwidth}
    \centering
    \begin{tabular}{@{}lcccc@{}}
      \toprule
      Model & Val acc.\ & F1 score & Optim.\ & \#param (M) \\
      \midrule
      ResNet18 – \textit{original paper implemented~\cite{zhixin2024chd}} 
        & 80.72\% & 80.11\% & SGD  & 11.7 \\
      \addlinespace[0.75ex]
      Inception v3 
        & 77.11\% & 71.73\% & Adam & 27.2 \\
      \addlinespace[0.75ex]
      ViT 
        & 63.86\% & 62.68\% & Adam &  5.7 \\
      \addlinespace[0.75ex]
      SWIN Transformer
        & 80.12\% & 79.51\% & Adam & 28.3 \\
      \bottomrule
    \end{tabular}
    \caption{Comparison of validation accuracy, F1 score, optimizer and parameter count for each model on DICOM dataset.}
    \label{tab:dicom-results}
  \end{subtable}
  \hfill
  \begin{subtable}[t]{0.38\textwidth}
    \centering
    \begin{tabular}{@{}lccc@{}}
      \toprule
      Model       & STFT              & Mel               & GAF               \\
      \midrule
      ResNet      & \cellcolor[HTML]{9AFF99}0.739 & 0.511             & 0.563             \\
      \addlinespace[0.5ex]
      EfficientNet& 0.711             & 0.563             & 0.511             \\
      \addlinespace[0.5ex]
      Inception   & 0.711             & \cellcolor[HTML]{9AFF99}0.620 & 0.563             \\
      \bottomrule
    \end{tabular}
    \caption{The Accuracy Results of Each Model and Data Representation on ZCHSound.}
    \label{tab:zch_results}
  \end{subtable}
  \caption{Model results.}
  \label{tab:merged}
\end{table*}

\begin{figure}[ht] 
  \centerline{\includegraphics[width=0.49\textwidth]{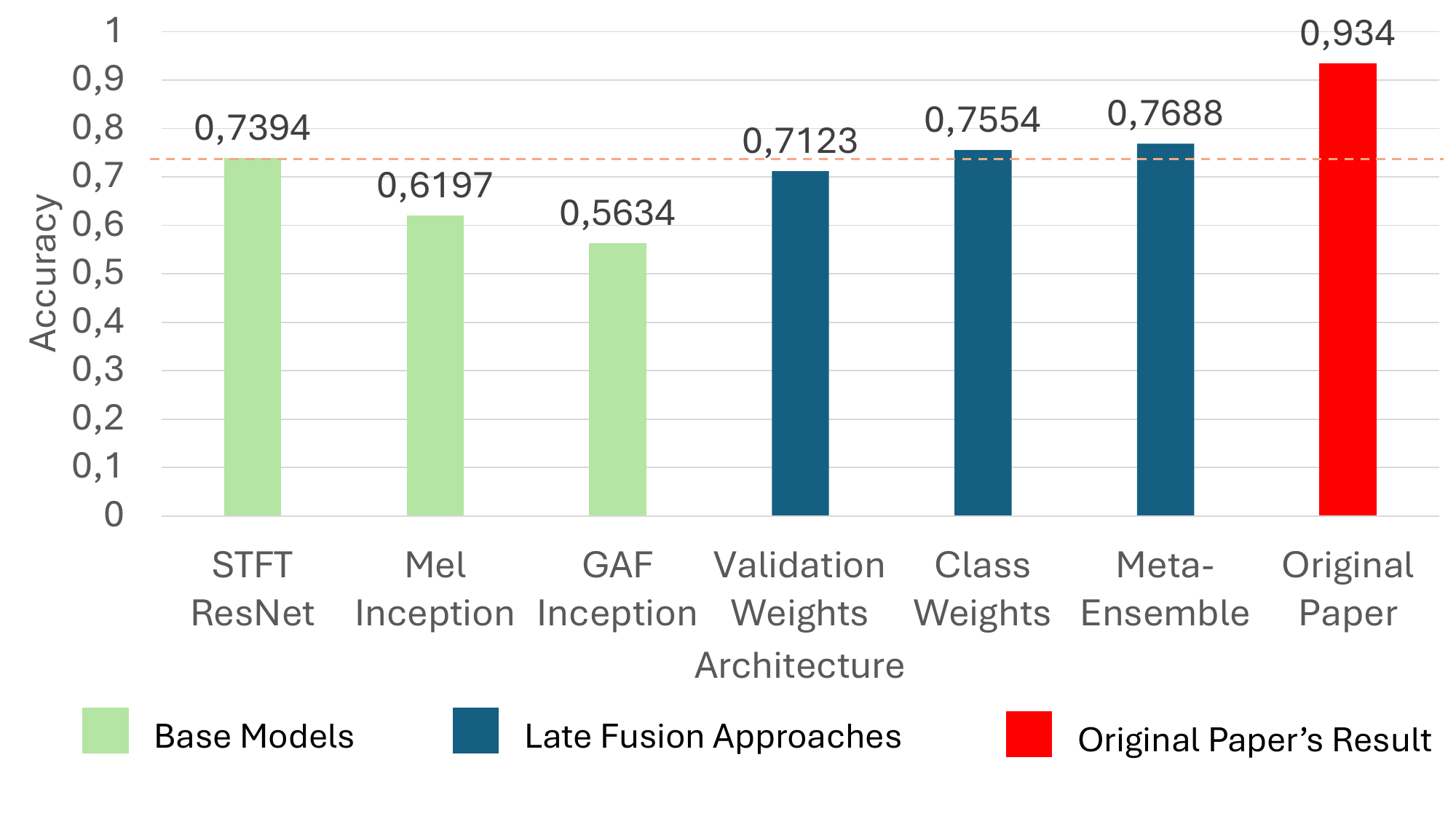}}
  \caption{Test Accuracy Results of Selected Base Model and Late Fusion Techniques.}
  \label{fig:audio_ensemble}
  \vskip-3mm
\end{figure}

After the models are selected, it is possible to implement the late fusion techniques. 
Figure \ref{fig:audio_ensemble} shows the test accuracy results of the selected base models and late fusion techniques. 
The results show that the approach with the validation accuracy weights has not improved the results. 
However, the class weights and the meta-ensemble techniques actually outperformed the individual base models, showing a 1.6\% and 2.7\% improvement, respectively. 
Still, these results are worse than what was reported by the original paper. \par

\begin{figure}[ht] 
  \centerline{\includegraphics[width=0.39\textwidth]{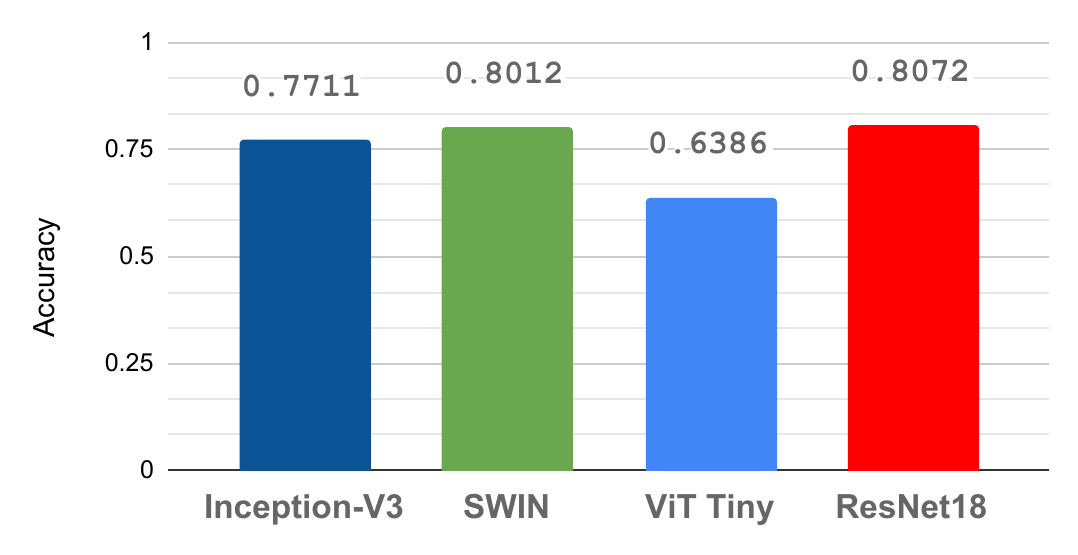}}
  \caption{Test Accuracy Results of Models on DICOM dataset.}
  \label{fig:dicom-res}
  \vskip-3mm
\end{figure}

From these results, we can see that STFT is the most suitable representation for the CHD classification, achieving at least 71\% accuracy.
While Mel and GAF representations show results better than the random guessing (20\% in this case), they are still low compared to STFT. 
This means they could potentially be replaced with other representations, such as Chirplet transforms as 2D images, or completely different modalities, such as original audio recordings or extracted descriptors. \par
Additionally, the selected architectures could be further tuned, both CNNs and the selected model for meta-ensemble. 
These changes could potentially improve the prediction results. \par

\subsection{DICOM image classification}
Table \ref{tab:dicom-results} shows the performance of four deep learning models such as ResNet-18, InceptionV3, ViT-Tiny, and Swin-Tiny on the preprocessed and augmented DICOM chest X-ray dataset. The primary reason of choosing the Transformer models comes from our observation that, to the best of our knowledge, there is no papers that applied SOTA ViT models on this problem. Hence, it can be said that, to some degree, it is the novelty of this paper. After preprocessing all the images we proceeded to training the models (details of the training phase are described in Methodology). 

The comparative analysis was performed using metrics like validation accuracy, F1-score and number of trainable parameters.\par
The ResNet-18 and Swin-Tiny models achieved over 80\% validation accuracy and Inception V3 performed well with the 77.11\% accuracy, while ViT-Tiny underperformed significantly in both metrics, which is illustrated in Figure \ref{fig:dicom-res}. \par

Best: ResNet-18 with almost 2x less parameters (i.e. complexity) than other two competing models (SWIN, Inception) of 11.7M was able to achieve the best accuracy of 80.72\% and F1 of 80.11\%. Though, it should be mentioned that it required 2x more epochs to converge (epochs = 43). It used residual connections to carry out deep training without vanishing gradients, and optimization with SGD allowed smoother convergence and better generalization. The use of this setup was very effective in learning the relevant spatial features from the preprocessed X-ray images.

Furthermore, Swin-Tiny also achieved a competitive performance (80.12\% accuracy) yet with 28.3M parameters. Though the hierarchical transformer structure with shifted windows was effective at capturing the multi-scale dependencies, the higher computational demand during training made this more costly.

General observations regarding Swin Transformer: 
\begin{itemize}
    \item VSD and Normal conditions are best identified with F1 scores of 0.83 and 0.82, respectively. Whereas ASD has the lowest F1 of 0.72. Possibly due to the class imbalance. ASD is has only 194 examples - lowest or it might be visually similar to other classes. For further analysis, Class Activation Maps may be needed.
    \item There was a little of overfit (around 20\% difference), even though a dropout layer was included.
\end{itemize}

The lowest performance was achieved by ViT Tiny, probably due to the limited number of parameters. Its patch based attention mechanism was not effective in extracting fine grained patterns from the Xray images.

Gaussian blurring and contrast enhancement had a major effect on all models. It resulted in more stable training and enhanced feature separation process as before applied techniques improved edge definition and normalized pixel intensity distributions which is shown in Histogram analysis.

Lastly, optimizer choice also played vital role in the implementation of the methodology. SGD trained ResNet-18 performed better than the other models that used Adam. However, SGD promoted more controlled learning dynamics that may have been the better choice on this relatively small and imbalanced dataset that Adam offered faster convergence, which may have introduced instability.\par

\section{Conclusion}

This project investigated the classification of CHD with several 2D representations of audio data and late fusion learning approaches. 
The results show that STFT representation is the most effective approach out of the selected ones, achieving 71\% accuracy with ResNet50-v2. 
Mel and GAF representation results were lower, suggesting that they may not capture the features required for CHD classification. \par
The late fusion approach based on the validation accuracy did not outperform the base model results, probably due to the dominance of a single model.
Class weight and meta-ensemble based approaches showed improvements of 1.6\% and 2.7\% respectively, confirming their efficiency in improving diagnostic accuracy. \par
Overall, however, these results did not outperform the results of the original work, showing that further refinements are required. The future work could include using alternative data representations (both images and other modalities), as well as the optimization of CNN and meta-ensemble architectures. 

Furthermore, our preprocessing pipeline—Gaussian blurring, contrast enhancement and augmentation—yielded a 6–7\% accuracy gain on the DICOM CHD‑CXR dataset. ResNet‑18 achieved the best performance (80.7\% accuracy, 80.1\% F1), whereas ViT‑Tiny underperformed. This work highlights the promising results of state-of-the-art vision transformers in CHD detection domain. Not only does the paper compare the accuracy but also raises the importance of considering complexity of models: higher complexity does not guarantee better results and even in some cases may be detrimental for the performance and deployability of the model in specific scenarios (e.g. medical facilities with limited resources cannot afford high-performing GPUs to support very complex models). Future work may include using adaptive parameter tuning, more advanced and thorough preprocessing analysis, and lastly increasing the dataset size via synthetic or real-life examples.

\section{References}
\renewcommand{\refname}{} 
\bibliography{report}

\end{document}